# A Hierarchical Framework for Collaborative Artificial Intelligence

James L. Crowley, Univ. Grenoble Alpes, France

Joëlle Coutaz, Univ. Grenoble Alpes, France

Jasmin Grosinger, Örebro University, Sweden

Javier Vazquez-Salceda, Universitat Politècnica de Catalunya, Spain

Cecilio Angulo, Universitat Politècnica de Catalunya, Spain

Alberto Sanfeliu, Universitat Politècnica de Catalunya, Spain

Luca Iocchi, Sapienza University of Rome, Italy

Anthony G. Cohn, University of Leeds, UK

## Abstract

We propose a hierarchical framework for collaborative intelligent systems. This framework organizes research challenges based on the nature of the collaborative activity and the information that must be shared, with each level building on capabilities provided by lower levels. We review research paradigms at each level, with a description of classical engineering-based approaches and modern alternatives based on machine learning, illustrated with a running example using a hypothetical personal service robot. We discuss cross-cutting issues that occur at all levels, focusing on the problem of communicating and sharing comprehension, the role of explanation and the social nature of collaboration. We conclude with a summary of research challenges and a discussion of the potential for economic and societal impact provided by technologies that enhance human abilities and empower people and society through collaboration with Intelligent Systems.

## 1. Introduction

Collaboration is a process where two or more agents work together as partners to achieve a shared goal [1]. Collaboration is a key challenge for creating Artificial Intelligence (AI) technologies that enhance human capabilities and empower people and society. We expect Collaborative AI to serve as a catalyst for the maturation and integration of AI technologies, enabling novel applications with important potential for economic and societal impact.

Research in Collaborative AI spans more than 40 years, with theories and models proposed by different scientific communities. Many of the communities agree on similar concepts, but with differences in focus, terminology and experimental methods. An important challenge for Collaborative AI is to unify and build on results from these communities, aligning their conceptual foundations and terminologies and building on their results.

In this article, we show how viewing collaboration as a hierarchy of perception-action cycles can provide a framework that unifies a broad spectrum of techniques and research problems for collaborative systems. We describe classical engineered approaches for components at each level and discuss modern alternatives based on machine learning. We then discuss some of the more salient research challenges that must be addressed to further develop collaboration with Intelligent Systems.

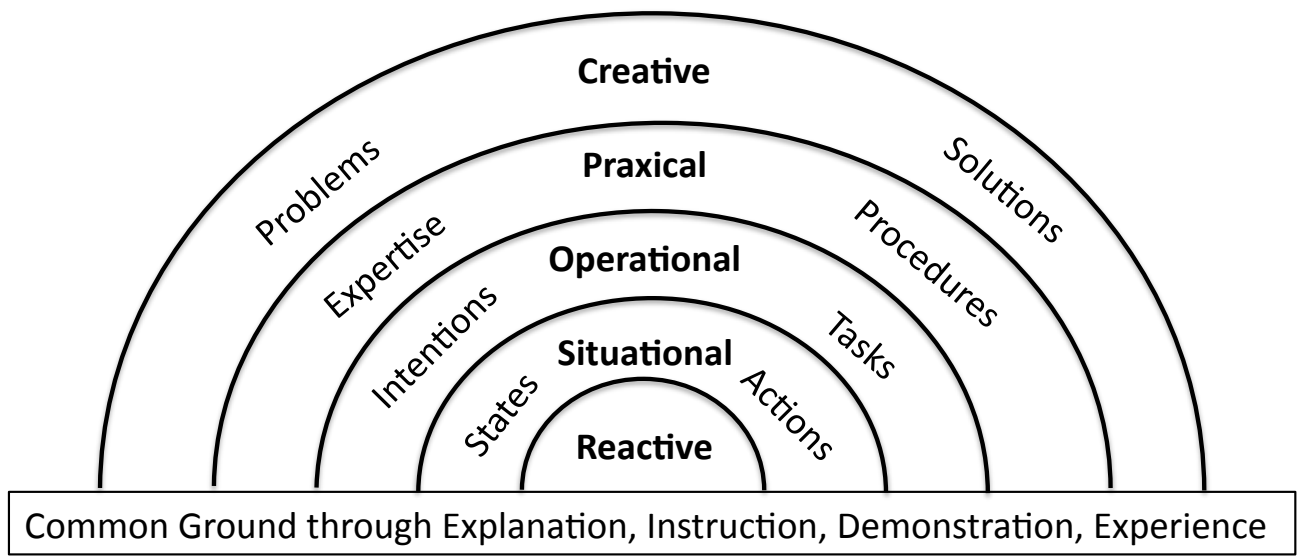

**Figure 1.** A hierarchy of capabilities for research on collaboration with intelligent systems. Collaboration at each level builds on abilities at the lower levels to determine solutions, procedures, tasks, and actions as shown on the right. Common ground for collaboration requires shared comprehension of situations (states), intentions, expertise and problems, as shown on the left. Common ground can be reached through explanation, instruction, demonstration and experience.

## 2. A Hierarchy of Perception-Action Cycles

Theories and experiments from Cognitive Science [2], Ergonomics [3] and Multi-modal Human-Computer Interaction [4], [5] show that humans observe, model, act and interact using multiple modalities over multiple temporal scales with multiple frames of references. Accordingly, human-AI collaboration can be organized as a hierarchy of perception-action cycles, each with specific representations for information. We refer to the levels in this hierarchy as reactive (sensori-motor), situational (spatio-temporal), operational (task-oriented), praxical (experience-based) and creative, as illustrated in Figure 1.

In the following, we provide definitions for each of these levels in terms of the information that can be represented and communicated. We summarize previous research on problems in each area and describe open research challenges, focusing on the problem of communicating and sharing comprehension, the power of explanations and the social nature of collaboration. We will use the example of a hypothetical personal service robot to illustrate collaboration at different levels. We will assume that our hypothetical robot is capable of indoor and outdoor navigation in the presence of pedestrians, with abilities to communicate, sense and express emotions and to interact socially with people using speech, gesture, vision and natural language. Enabling technologies for such robots are the subject of active research, including within the laboratories of the authors of this article.

**2.1 Reactive Collaboration**

Reactive collaboration assumes a form of tightly-coupled interaction where the actions of each agent are immediately sensed and used to trigger actions by the other. Reactive collaboration with humans requires that machines sense and act with a similar time-scale as humans. Sensori-motor reflexes in humans occur over a time scale of 80 to 300 milliseconds with reaction times determined by the number of neural layers between the sensing organ and the muscle activation units. Effective interaction requires that the temporal and physical properties of the machine be tuned to the sensori-motor reflexes of the human collaborator. For example, a computer mouse must immediately display movements of a cursor on a computer screen within the perceptual time scale of the human user [6]. Even a small lag in display can have serious consequences for usability.

For our personal robot, an example of reactive collaboration is provided by navigation in a crowd of moving pedestrians. The potential for collisions can be estimated from the rate of change of bearing angle to pedestrians or obstacles. A small rate of change of bearing, combined with a negative change in distance indicates a potential for collision. Time to collision can be estimated by the relative distance divided by the rate of change of distance. Normally, collision can be avoided reactively by adjusting velocity or heading so as to increase the rate of change in bearing or increase the time to contact. For converging pedestrians, there is a social aspect to collision avoidance, with one pedestrians yielding to the other, or both pedestrians adapting their paths so as to avoid collision [7]. Integrating this social dimension into robot collision avoidance requires monitoring the motion of the pedestrians, predicting intention and reacting in a compatible manner.

Classically, such systems have been implemented by directly coupling sensor output to motor control at design time. Techniques for learning reactive control of behaviors at run-time have recently been demonstrated. End-to-end systems can be trained using neural networks that transform perceptual signals directly to motor commands [8], providing a new technology for Demonstration Learning [9]. Reinforcement Learning can be used with Deep Learning [10] to learn control policies for mapping perception to action from experience. Deep Reinforcement Learning is increasingly used to learn robot behaviors or skills as networks that directly transform perception to action. Such techniques can be used to pre-learn fixed behaviors for interaction. However, Deep Reinforcement Learning can require an extremely large number of trials, typically performed offline using a simulator. Effective and efficient on-line learning through interaction with a teacher remains an open challenge.

**2.2 Situational Collaboration**

Situational collaboration refers to an interaction where perception and action are mediated by shared awareness of situation. Situation awareness has long been recognized as a core competence for intelligent behavior, as well as survival in critical environments. The term can be traced to the early 20th century, where situation awareness was identified as a crucial skill for crews in military aircraft. Situation awareness has been recognized as a foundation for successful decision-making across a broad range of domains.

Factors related to situation awareness are a core concern for ergonomics. According to Endsley [3], situation awareness is the perception of environmental elements and events with respect to time or space, the comprehension of their meaning, and the projection of their future status. Such abilities may be modeled as a process of construction, maintenance and use of a situation model.

A situation model is a state-space model composed of four types of information [2]:

1. A spatio-temporal framework (spatial locations, time frames)
2. Entities (people, objects, ideas, etc.)
3. Properties of entities (color, size, shape, emotions, etc.)
4. Relations (spatial, temporal, causal, ownership, family, social, etc.)

For mobile robot navigation, a classic example of a situation model is the network of places [11]. Each place in the network is a state defined by a name, and the proximity of the robot to a world position. Each place can be associated with robot behaviors using rules or behavior trees [12]. Situation models decouple perception and cognition from behavior, providing the basis for flexible systems that can assimilate observations, predict consequences, and determine appropriate actions for services using adaptive behaviors that can be learned and refined.

Situational collaboration requires that people and systems share a situation model, including a common vocabulary for entities, relations and situations, and the actions that result in changes of situation. Established approaches for machine perception use supervised learning to pre-train object detectors with labeled training data from a pre-defined set of categories. This approach limits perception to a closed set of pre-defined entities and relations, thus limiting the set of situations that can be modeled or communicated. An important challenge for situational collaboration is to provide a means to perceive and communicate information about new entities and new relations. This requires an ability to learn from demonstration, explanation and experience.

The ability for an intelligent system to generate or comprehend an explanation about a current or intended situation remains a fundamental challenge for collaborative interaction. Sharing comprehension of a situation through explanation is an important challenge for collaboration with intelligent systems that is beyond the current state of the art.

## 2.3 Operational Collaboration

The operational level concerns the planning and execution of tasks. Information at the operational level includes the current and desired situations, their expression as intentions, goals and sub-goals, tasks and sub-tasks, and plans of actions that can be used to attain the desired situation. This level can also concern actions that can be used to attain or maintain a stable situation as well as detection of threats and opportunities.

Operational collaboration requires sharing authority. Authority may be shared with a strict hierarchy, where one agent has the power to over-rule the actions of the other as with an aircraft or maritime crew. Authority is often shared using a protocol where each agent has a primary authority over a particular task domain, with a possibility of accepting delegation of authority in other domains. Authority may also be shared equally where each agent is free to initiate tasks according to its understanding of the common goals and current situation, as can occur in some forms of team sports such as football or ice hockey. Operational collaboration can be facilitated by a shared comprehension of the current situation, desired goal, and plans of actions that can be used to attain the goal.

In the case of our personal service robot, it is normally assumed that the robot's patron has authority over the robot. In this case, the patron would ideally delegate responsibility to the robot for certain tasks such as cleaning or cooking. For example, the patron may ask the robot to monitor sauce on a stove and make sure that it does not burn. Alternatively, it may be desirable for the robot to have the authority to proactively initiate tasks to assist the human or avoid an undesirable situation.

Several scientific and engineering domains have proposed techniques for solving problems at the operational level. In classic AI, problem solving is performed in a state space [13], with an initial state, a set of desired goal states and a set of actions that can be used to move between states. Classical AI planning techniques are concerned with how to represent the problem space and how to use the space to automatically develop plans of actions. Task planning [14] has been extended to the more general problem of developing a possibly parallel sequence of operations that may be performed by multiple agents in a coordinated manner to attain a goal. In such cases, a library of planning algorithms and behaviors (perception-action reflexes) can be used to provide approximate solutions. However, with such approaches, the problem space is based on a closed, pre-defined set of entities, and associated plans and behaviors are programmed in advance. In many real-world domains, such approaches are made impractical by the prohibitively large size of the state space and the cost of acquiring information.

A framework for the study of operational collaboration has been provided by Activity Theory, a research framework that originated in Soviet psychology in the 1920s [15]. Activity Theory holds that the constituents of an activity are not fixed but can dynamically change as conditions (situations) change. As actions are repeated, they may be compiled, or "operationalized" into operations. When confronted with a failure or an unfamiliar situation, operations may be decompiled into explicit plans for re-planning to overcome the difficulty.

Activity Theory is essentially a framework for description and analysis. An attempt to extend this framework to a predictive theory is provided by Situated Action Theory. Suchman [16] observes that human action is constantly constructed and reconstructed from dynamic interactions with the material and social worlds. Situated Action Theory emphasizes the importance of the environment as an integral part of the cognitive process. The subject of study is not the individual or the environment, but the interaction between the two. Situated action models emphasize the emergent, contingent nature of activity, and stress that activity emerges from the particularities of a situation.

An alternative form of activity theory is provided by proactive systems. Proactive behavior emerges from the particularities of the current situation and possible future situations, including the individual and the environment [17]. Such behavior is increasingly recognized as important for learning to interact as well as for generating a sense of affection and trust [18].

An ability to comprehend a human explanation for a plan of action, including authorizations or limits to authority remains an important challenge for collaboration with intelligent systems. Operational collaboration can benefit from exchanges of information about procedures and interaction protocols based on experience or training. This is provided by the praxical level.

**2.4 Praxical Collaboration**

Praxical collaboration involves the exchange of knowledge about actions and procedures for collaboration based on experience or training. Praxis is knowledge about how to act. The term "Praxical" was proposed by Heidegger [19] to refer to the knowledge acquired through experience of how to use tools. Heidegger proposed a view of learning where capabilities are acquired through practical experiments. For example, to learn about a hammer, it is necessary to hammer things and observe the effects. Here we are extending the term Praxical to include the exchange of information about the concepts, actions and techniques, whether preprogrammed, learned from theory or training, or acquired from experience. Information represented at this level is about how to perceive and modify the environment, how to predict phenomena including the effects of action, and how to develop plans, select behaviors and perform actions. Praxis is particularly important for social interaction, as societies, cultures, and local populations continually evolve social practices. Learning proper protocols is fundamental to collaboration.

Praxical collaboration requires protocols for how to exchange information and work together. Much of everyday human interaction is guided by social norms that provide protocols for interaction. The capability to recognize a social situation and adopt the appropriate social role is fundamental to human interaction at this level. Socially collaborative machines require abilities to perceive and recognize social situations, and to behave in a manner that complies with the roles and interaction protocols dictated by social norms. While such capabilities can be learned through experience, human learning is generally greatly facilitated by explanations from a parent or teacher. People acquire knowledge of social norms from explanation and training refined by experience. Learning from explanations is an important open challenge.

In the case of our personal service robot, the robot must be able to adapt to the individual preferences and routines of the patron in order to be accepted. Ideally, this would require an ability to interpret and learn from explanations about how objects should be found or stored and instructions for how tasks should be performed. It would also require that the system has an ability to generate explanations for problems encountered while performing tasks in order to request assistance or to justify its decisions and actions. A key challenge at this level is the capability to generate and comprehend explanations about individual preferences.

An even more challenging form of praxical collaboration is learning to act so as to please a partner. An elderly care robot that shares a living space with a person must rapidly adopt behaviors that please and evoke affection. Learning to behave so as not to annoy, and learning to interact so as to stimulate positive and healthy emotions will be essential both to acceptance and to success as a resource for elderly care.

Current technologies for social interaction with intelligent systems generally rely on preprogrammed dialogs and rules with little or no ability to adapt to local customs or individual variations. An important challenge in this area is to develop a technology for systems that can acquire and refine praxical abilities for protocols for socially correct interaction through training, explanation and experience, balancing pre-programmed abilities with learned behaviors.

**2.5 Creative Collaboration**

Creative collaboration refers to a form of interaction where two or more partners work together to solve a problem or create an original artifact. In the most effective forms of creative collaboration, each partner evaluates the comprehension and analysis of other partners in order to offer constructive criticism or to reinforce and build on emerging insights. When two or more partners work well together, a form of creative resonance emerges in which each partner improves and builds on the ideas of the others.

A classic example of creative collaboration with an intelligent system was provided by the R1 expert system used by Digital Equipment Corporation (DEC) to configure Vax computers in the 1980s [20]. R1 was a rule-based expert system that used approximately 500 forward chaining rules to empower a salesperson to assemble a compatible configuration of software and hardware. The salesperson would describe customer requirements to the R1 system, and use R1 to configure a working system. This approach dramatically improved the product acceptance rate for Vax computers, with 10-fold payback on the initial investment within the first year of use. In the case of a personal robot, creative collaboration would be useful for tasks such as developing a common strategy to search for a lost object, or inventing a new recipe to accommodate missing ingredients.

Creative collaboration can be expected to find widespread application in professional areas including systems design, discovery and artistic creation. Expert systems technologies failed to achieve widespread use, largely because of the very high cost of encoding human expertise and the fact that many areas of human expertise require praxical knowledge that could not be encoded with the formal methods of early AI. The emergence of powerful new machine learning techniques that can acquire reactive, situational, operational and praxical knowledge offers the possibility of building intelligent systems that can amplify human capabilities through creative collaboration. For example, an intelligent system for collaborative programming has recently become available using Open-AI's Codex system based on GPT-3 [21] and many more such services can be expected in the near future. The challenge is how to apply these new learning technologies to build intelligent systems that function as effective collaborators to empower people.

# 3. Core Research Challenges for Collaborative AI

A compilation of the open research challenges posed by collaborative AI resulted in a long list of domain specific research problems. However, grouping these problems according to the framework level revealed a repeated pattern based on three core abilities: Comprehension, Explanation and Learning, as shown in figure 2.

**3.1 Comprehension**

Comprehension can be defined as the ability to understand the meaning and importance of a sensory perception or of a linguistic construct. In common language, the terms comprehension and understanding are often used interchangeably. However, understanding tends to be overloaded with different meanings, whereas comprehension has well defined technical definitions in the educational and cognitive sciences. For example, reading comprehension is the ability to process a text, understand its meaning, and to integrate this meaning with what the reader already knows [22]. Using the terminology of ergonomics [3], comprehension can be summarized as a process of assimilation, association, and anticipation. A sensory percept is expressed in an internal representation (assimilation) and associated with various forms of memory. The meaning of the percept emerges from such associations, allowing the agent to predict or anticipate consequences and determine actions.

An ability to comprehend is fundamental for collaborative intelligent systems. At the reactive level, the agent must be able to comprehend perceptions and the effects of actions in order to provide reactive control, as well as maintain an accurate model of a situation. For situational and operational collaboration, a system must be able to comprehend explanations of the current and desired situations and to generate explanations of its own comprehension to share with a partner. Comprehension and explanation emerge as fundamental to interaction at all levels.

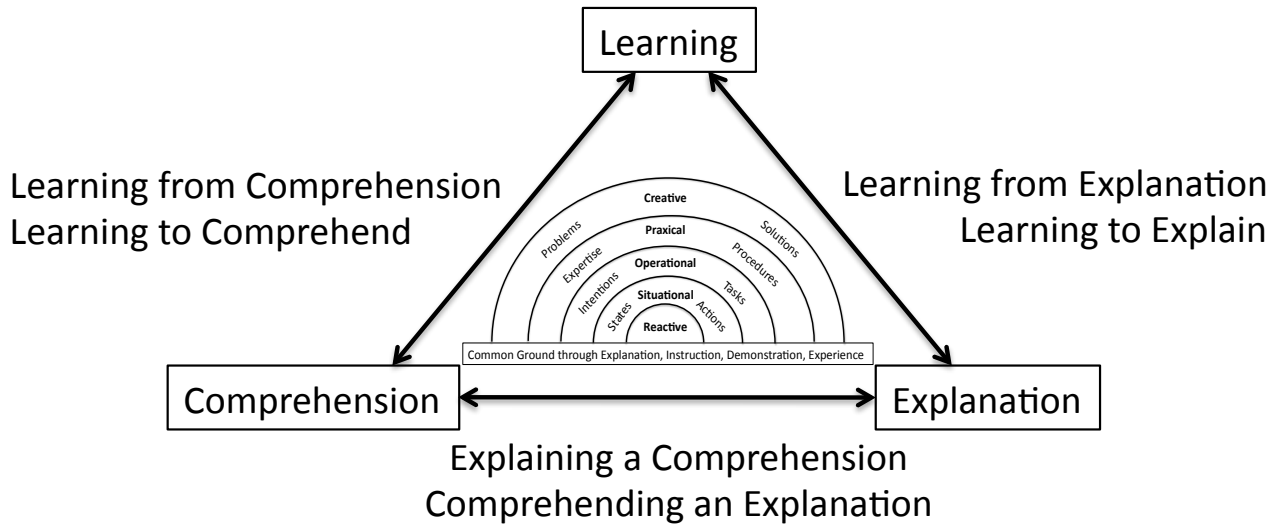

**Figure 2.** A technology for Collaborative Intelligent Systems requires abilities for comprehension, explanation and learning at each level of collaboration, with abilities at each level building on the abilities of lower levels. The research challenges associated with comprehension, explanation and learning at each level require different technical expertise, depending on the nature of the information that is processed at each level.

**3.2 The Role of Explanation**

In a recent article, John Carroll argues that a proper technology for "Explanatory AI" is fundamental for trust between humans and AI [23]. Explanatory AI explains itself, providing an account of why it does what it does. However, such explanations are highly subjective and deeply contextual [24], with needs and expectations depending on the degree of domain and technical expertise of the human partner.

An explanation can be defined as a statement that makes something clear or justifies an action. For situation modeling, the "something" is the situation, including the underlying entities and relations as well as associated actions and intentions. For operational collaboration, an explanation can provide a description of the sequence of intended actions that can take a situation to a desired state, as well as a description of the sequence of intermediate situations that can be used to ensure the proper execution of actions and operations. For praxical collaboration, explanations can be used to share knowledge about how to obtain information and coordinate actions based on habits and social norms. Explanations can also facilitate agreement on protocols for interaction and collaboration, facilitating coordinated action and recognition of intention. Explanations can be used to describe hypotheses for creative collaboration. An ability to generate and interpret explanations is key for all levels of collaboration with intelligent systems.

Explanations can be structured as narratives that provide answers to the Quintilian questions Who, What, Why, When, Where, and How. Specifying these elements is key to establishing a shared situation model and a shared agreement on operational plans and authority, whether describing past, present or future situations. "What" and "where" describe the entities and relations that compose a situation. "Why" describes the desired or goal situations. "How" concerns the sequence of actions or operations that can be used to reach the goal. "When" describes the conditions under which the actions and operations can be performed. "Who" assigns the operational authority to perform the actions, or establishes a protocol for determining authority during the operation.

Such a narrative structure can substitute for a lack of experience by providing grounding for interpreting instructions. Explanations can be used to compensate for the lack of experience when providing instructions. They can also be used to diagnose and learn from the results of operations "after the fact" when operations fail to provide a desired outcome. An explanation can help identify whether the failure was due to incomplete or inaccurate model of the situation or the result of erroneous assumptions or some other cause. A narrative that explains the understanding of the situation and the reasons for selecting actions can be used to learn from the failure and improve operations for the future.

Intelligent collaboration requires abilities to share comprehension. Explanations provide a powerful technique for sharing situation models, operational plans, praxical knowledge and creative solutions. Developing technologies for dynamically generating explanations and for interpreting explanations are an important challenge for collaboration with intelligence systems. In addition to comprehension and explanation, collaborative intelligent systems require an ability to learn.

### 3.3 The Challenge of Learning

Learning can be defined as the acquisition of new abilities through explanation or experience. An ability to learn is even more fundamental than comprehension or explanation. The enormous complexity of the real world imposes a requirement to continue to acquire and adapt abilities. Indeed, most definitions of intelligence include an ability to learn as a core ability.

Note that learning is distinct from comprehension. Comprehension refers to acquisition and maintenance of a description for controlling a process, modeling a situation, planning an operation, or agreeing methods to achieve goals. Learning concerns acquisition of abilities: the means or capacity to perceive, comprehend, and act.

Learning to recognize entities can be particularly challenging for functional categories of objects based on affordances. Affordances are the qualities or properties of an object that define possible uses for the object and indicate how it can be used [25]. Functional categories are sets of entities that share affordances. For humans, learning to recognize and name functional categories is facilitated by experience. Without reference to the experience of sitting, machines are reduced to superficial detection of chairs based on shape and appearance.

Advances in machine learning have recently provided an enabling technology for systems that can learn through observation and interaction. Transformers and self-attention [26] have become the dominant approach for natural language processing (NLP) with systems such as BERT [27] and GPT-3 [20] rapidly displacing more established Recurrent and Convolutional Network structures with an architecture composed of stacked encoder-decoder modules using self-attention. This provides an important new technology that appears well-suited for linguistic communications at all levels, including sharing information about situations, agreeing on tasks at the operational level, as well as sharing praxical knowledge and communicating for creative collaboration. Applying such technology for collaboration with intelligent systems raises many new and interesting research challenges.

### 3.4 Finding Common Ground

In a recent paper in Nature [28] the authors argue for a science of cooperative intelligence based on machine abilities to understand, communicate and interact with people, under the guidance of norms and institutions. The authors referred to this as "finding common ground" with AI systems. Finding common ground and the related problem of mutual understanding provide an interesting perspective on research challenges for collaborative AI. This requires abilities to comprehend, explain and learn at reactive, situational, operational, praxical and creative levels raising challenging problems for a variety of scientific disciplines. Differences in the expertise required to address problems at each level result from the nature of the information expressed: sensori-motor signals, symbolic entities and relations, plans of action, or general knowledge about how to accomplish tasks and create new artifacts. The study of Collaborative Artificial Intelligence is not the exclusive domain of any one scientific community.

## 4. Discussion and Conclusions

We have described a hierarchical framework for collaborative intelligent systems. Each level of the hierarchy concerns interaction with distinct forms of information: Sensori-motor signals for the reactive level, entities and relations for the situational level, tasks and plans for the operational level, domain specific knowledge about how to perceive and act for the praxical level, and problems, hypotheses and solutions at the creative level.

This framework has been designed to provide a problem space for research, grouping related challenges into subcategories according to the information that is processed and the nature of the interaction. Such a problem space facilitates formulation and comparative evaluation of competing techniques. However this framework can also serve as a reference model for designing systems, providing a functional decomposition for collaborative intelligent systems.

All levels are concerned with natural language communications. Reading, writing, listening and speaking have substantial sensori-motor components. Much of natural language communication is about describing situations. For example, the verb of a sentence is a predicate describing the relation between the subject entity and one or more object entities. Natural language can be used for coordinating actions during operational collaboration. Natural language is highly effective and widely used for communicating praxical knowledge and for creative collaboration.

Enabling technologies for collaboration with intelligent systems have potential for significant societal impact and wealth generation. An obvious example is in computer games and virtual worlds (the metaverse) where enabling virtual characters with abilities for situation understanding, operational collaboration and creative problem solving can have enormous impact. Similarly, virtual musical groups with simulated players that can sense and react musically to a musician can engender creativity by empowering musicians to explore new forms of music without the clash of egos that can occur in real musical bands.

Chatbots are another area where an enabling technology for collaborative problem solving could provide important wealth generation. Most current chatbot technologies rely on pre-programmed linguistic patterns to generate plausible natural language responses to queries, while restricting responses to preprogrammed answers or interpretations of results from search engines. Empowering chatbots with an ability to creatively explore solutions with users would have an enormous appeal, for example by providing virtual customer service agents for online commerce that could guide and assist users in planning purchases.

Technologies to permit humans and intelligent systems to collaboratively analyze problems and determine solutions augmented with explanations and courses of possible actions can have enormous impact for social impact and scientific understanding. Finding common ground at all levels is key.

**Acknowledgements**

This work has been partially supported by the MIAI Multidisciplinary AI Institute at the Univ. Grenoble Alpes (MIAI@Grenoble Alpes - ANR-19-P3IA-0003), by the EU H2020 ICT AI4EU (grant agreement No. 825619) and by the EU H2020 project Humane AI Net (grant agreement No. 952026).

# Bibliography

1. L. G. Terveen, An Overview of Human-Computer Collaboration. *Knowledge-Based Systems*, 1995.
2. P. N. Johnson-Laird, *Mental Models*, MIT Press, Cambridge, MA, USA, 1989.
3. M. R. Endsley and D. J. Garland (Eds.). *Situation Awareness Analysis and Measurement*. Lawrence Erlbaum, Mahwah, NJ, 2000.
4. M. Turk, Multimodal interaction: A review, *Pattern Recognition Letters*, 36, pp.189-195, 2014.
5. S. Oviatt, B. Schuller, P. Cohen, D. Sonntag, and G. Potamianos, *The Handbook of Multimodal-Multisensor Interfaces: Foundations, User Modeling, and Common Modality Combinations*, Morgan and Claypool, 2017.
6. S.K. Card, T.P Moran, and A. Newell, 1983. *The psychology of human-computer interaction*. Lawrence Erlbaum Associates. Hillsdale New Jersey, 1983.
7. T. Kruse, A. K. Pandey, R. Alami, and A. Kirsch, "Human-Aware Robot Navigation: a Survey", in *Robotics and Autonomous Systems*, Elsevier, vol. 61, no. 12, pp.1726–1743, 2013.
8. S. Levine, C. Finn, T. Darrell and P. Abbeel, "End-to-End Training of Deep Visuomotor Policies" in *The Journal of Machine Learning Research*, vol. 17, no. 1, pp. 1334–1373, 2016.
9. B. D. Argall, S. Chernova, M. Veloso and B. Browning, "A Survey of Robot Learning from Demonstration", in. *Robotics and autonomous systems*, vol. 57, no. 5, pp. 469–483, 2009.
10. V. Mnih et al., "Human-level Control through Deep Reinforcement Learning" in *Nature*, *518*(7540), pp. 529–533, 2015.
11. J. L. Crowley, "Navigation for an intelligent mobile robot", *IEEE Transactions on Robotics and Automation*, vol. 1 no 1, pp. 31-41, 1985.
12. M. Colledanchise and P. Ogren, "How Behavior Trees Modularize Robustness and Safety in Hybrid Systems", in *IEEE/RSJ Int. Conf. on Intelligent Robots and System,* IROS-2014, pp. 1482–1488, Sept 2014.
13. A. Newell and H. A. Simon, *Human problem solving*, vol. 104, no. 9, Prentice-hall, Englewood Cliffs, NJ, 1972.
14. M. Cirillo, L. Karlsson and A. Saffiotti, "Human-aware Task Planning: an Application to Mobile Robots", in *ACM Transactions on Intelligent Systems and Technology* (TIST), vol. 1, no. 2, pp.1–26, 2010.
15. B. Nardi, *Context and Consciousness*: *Activity Theory and Human Computer Interaction*, MIT Press, Cambridge Mass, 1996.
16. L. Suchman, *Plans and Situated Actions: The Problem of Human-Machine Communication*, Xerox, Palo Alto Research Centers, 1987.
17. H. Harman and P. Simoens, "Action Graphs for Proactive Robot Assistance in Smart Environments", in *Journal of Ambient Intelligence and Smart Environments*, vol. 12, no. 2, pp. 79–99, 2020.
18. J. Grosinger, F. Pecora, and A. Saffiotti. "Robots that maintain equilibrium: Proactivity by reasoning about user intentions and preferences." *Pattern Recognition Letters* 118, pp85-93, 2019
19. B. Bolt, *Heidegger Reframed*, Bloomsbury Publishing, 2010.
20. J. McDermott, "R1: A Rule-Based Configurer of Computer Systems", in *Artificial intelligence*, vol. 19, no. 1, pp. 39–88, 1982.
21. T. B. Brown, B. Mann, N. Ryder, M. Subbiah, J. Kaplan, P. Dhariwal, and D. Amodei, Language models are few-shot learners, *Advances in neural information processing systems*, vol. 33, pp. 1877-1901, 2020
22. W. Kintsch, , *Comprehension: A paradigm for cognition*. Cambridge University Press, 1998.
23. J. M. Carroll, "Why should humans trust AI?", *Interactions*, vol. 29, no. 4 , pp73-77, ACM, July-August 2022.
24. J. Coutaz, J. L. Crowley, Dobson, S., and Garlan, D., Context is key. *Communications of the ACM*, *48*(3), pp49-53, 2005.

25. J. J. Gibson, *The theory of affordances*, Lawrence Earlbaum Associates, Hillsdale, NJ., pp. 67–82, 1977.
26. A. Vaswani, N. Shazeer, N. Parmar, J. Uszkoreit, L. Jones, A. N. Gomez, Ł. Kaiser and I. Polosukhin, "Attention is all You Need, *Advances in neural information processing systems*, 2017, vol. 30*,* 2017.
27. J. Devlin, M.-W. Chang, K. Lee, and K. Toutanova, BERT: Pre-training of Deep Bidirectional Transformers for Language Understanding. In *Proceedings of the 2019 Conference of the North American Chapter of the Association for Computational Linguistics*, Vol 1, pp 4171–4186, Minneapolis, Minnesota. 2019.
28. A. Dafoe, Y. Bachrach, G. Hadfield, E. Horvitz, K. Larson, and T. Graepel, "Cooperative AI: Machines Must Learn to Find Common Ground", in *Nature*, vol. 593, pp. 33–36, May, 2021.

**James L. Crowley** is Professor Emeritus at the Institut Polytechnique de Grenoble, and holds the Chair of Collaborative Intelligent Systems at the Univ. Grenoble Alpes Multidisciplinary Institute for Artificial Intelligence. His current research combines multi-modal interaction with cognitive modeling to explore new forms of interaction with intelligent systems. Contact him at James.Crowley@grenoble-inp.fr.

**Joëlle Coutaz** is an Honorary Professor at the Univ. Grenoble Alpes and founder of the Engineering Human–Computer Interaction group within the Grenoble Informatics Laboratory (LIG). Her research interests include Human-Computer Interaction, multimodal and tangible interaction, user interface plasticity, and end-user development for smart homes and ubiquitous computing. Contact her at joelle.coutaz@imag.fr.

**Jasmin Grosinger** is a post-doctoral researcher in the Cognitive Robotics Lab at the Center for Applied Autonomous Systems (AASS) at Örebro University, Sweden. Her research interests research concern AI for autonomous agents in a real-world setting. Contact her at Jasmin.Grosinger@oru.se.

**Javier Vázquez-Salceda** is Associate Professor at the Universitat Politècnica de Catalunya. His research is focused on the design and development of social frameworks to ease the interaction among software and/or physical agents, especially in distributed applications mixing humans and AI systems for complex domains such as Smart Cities, eGovernment, eCommerce or Medicine. Contact him at jvazquez@cs.upc.edu.

**Cecilio Angulo** is Full Professor at the Universitat Politècnica de Catalunya and founder of the Intelligent Data Science and Artificial Intelligence Research Center. His research interests include social and cognitive robotics, reinforcement learning and human-robot interaction. Contact him at cecilio.angulo@upc.edu

**Alberto Sanfeliu** is Professor of Computational Sciences and Artificial Intelligence at the Universitat Politècnica de Catalunya. He is coordinator of the research Artificial Vision and Intelligent System Group and head of the Mobile Robotics group. His research interest includes Intelligent Robotics, Human-Robot Interaction, Computer Vision and Pattern Recognition. Contact him at alberto.sanfeliu@upc.edu

**Luca Iocchi** received a PhD in Engineering in Computer Science and he is Full Professor at Sapienza University of Rome, Italy. His research interests are in the fields of in artificial intelligence and robotics, including cognitive robotics, task planning, multi-robot coordination, reinforcement learning, human–robot interaction, and social robotics. Contact him at: iocchi@diag.uniroma1.it

**Anthony G Cohn** is Professor of Automated Reasoning at the Univ. of Leeds, and a Fellow at the Alan Turing Institute in the UK. His research interests include knowledge representation and reasoning, in particular relating to spatial information and common sense, ontologies, decision support systems, cognitive vision, robotics, grounding language in vision and evaluation of AI systems. Contact him at a.g.cohn@leeds.ac.uk.